\documentclass[11pt,letterpaper]{article}

\usepackage[margin=1in]{geometry}
\usepackage[T1]{fontenc}
\usepackage{microtype}

\usepackage{mathtools,amssymb,amsthm}
\usepackage{newtxtext,newtxmath}

\newtheorem{theorem}{Theorem}[section]

\theoremstyle{definition}

\theoremstyle{remark}
\newtheorem{remark}[theorem]{Remark}

\usepackage{graphicx}
\usepackage{booktabs}
\usepackage{tabularx}
\usepackage{enumitem}

\setlist{
  itemsep=0.25em,
  topsep=0.4em
}

\usepackage[numbers,sort&compress]{natbib}

\usepackage{xcolor}
\definecolor{linkblue}{RGB}{25,65,120}

\usepackage[
  pdfusetitle,
  colorlinks=true,
  linkcolor=linkblue,
  citecolor=linkblue,
  urlcolor=linkblue
]{hyperref}

\newcommand{\ACz}{\mathsf{AC}^0}
\newcommand{\TCz}{\mathsf{TC}^0}

\newcommand{\LOGSPACE}{\mathsf{LOGSPACE}}
\newcommand{\POLYTIME}{\mathsf{PTIME}}
\newcommand{\NP}{\mathsf{NP}}
\newcommand{\UHAT}{\mathsf{UHAT}}
\newcommand{\AHAT}{\mathsf{AHAT}}
\newcommand{\SMAT}{\mathsf{SMAT}}

\newcommand{\MAJ}{\mathsf{MAJ}}

\title{\textbf{On the Expressive Power of Transformers}}

\author{
  Phokion G. Kolaitis\\
  \small University of California Santa Cruz\\
  \small Santa Cruz, CA\\
  \small \texttt{kolaitis@ucsc.edu}
  \and
  Rik Sengupta\\
  \small IBM Research\\
  \small Cambridge, MA\\
  \small \texttt{rik@ibm.com}
}

\date{August 2026}

\begin{document}

\maketitle

\begin{abstract}
Multi-layer transformers form the critical component of essentially all large language models (LLMs) in use today. Because of their ubiquity and computational capability, there is a rapidly growing  body of work that aims to precisely calibrate the expressive power of transformers as language recognizers by comparing them against standard models of computation  studied for decades by the theoretical computer science community. In this endeavor, circuit complexity has by and large emerged as the ``correct'' branch of computational complexity to analyze the expressive power of transformers; the reason is that parameterizing transformers by the various resources they use, such as  attention and precision, leads to direct comparisons with different classes of circuits parameterized by resources such as type of gates, size, and depth. Here, we present an overview of selected results that delineate the expressive power of transformers using  concepts and methods from circuit complexity.
\end{abstract}


\medskip


\tableofcontents

\section{Introduction}\label{sec:introduction}

Transformers have become the default computational substrate of modern large language models (LLMs), and yet their formal capabilities remain only partially understood at best. For researchers in logic and computational complexity, this presents a challenge and an opportunity: formalize  an architecture that has extraordinary empirical success in order to study it as a family of resource-bounded computation models and to  compare its power with classical hierarchies underlying  standard models of  computation. Indeed, a large body of recent literature has pursued this exact endeavor. By treating sequence length, depth, attention heads, numerical precision, positional encodings, and so on as explicit resources, one obtains variants of transformers  whose expressive power can be related to that of automata and circuit classes.

This brief survey focuses on transformers as language recognizers and their connections to circuit complexity. The underlying principle is that attention layers behave like structured, parallel stages of computation, which makes them comparable to bounded-depth circuits with restrictions on gates, fan-in, size, and uniformity. This perspective clarifies both what transformers can simulate and where limitations arise. It also exposes how small choices in setting up the architecture --- soft versus hard attention, fixed versus unbounded precision, and so on --- can significantly affect the expressive power. Our aim is to set up the formal architecture carefully, highlight
 a few of the results established so far, and
provide a flavor of some of the proof techniques. 

For a more detailed overview, we refer the reader to several other extensive surveys of the field, including ones on neural networks and formal languages \citep{survey-ackermancybenko}, RNNs and transformers \citep{survey-merrill}, transformer expressivity \citep{survey-strobl+}, and the transformer cookbook \citep{cookbook}.
\section{The Architecture}\label{sec:architecture}

\subsection{Transformers as Language Recognizers}\label{sec:recognizers}

Before  describing each of its components formally, we describe the transformer architecture informally, so that we establish how transformers can serve as language recognizers. A transformer can be viewed as a particular neural network consisting of an input layer, one or more hidden layers, and an output layer. The input to a transformer is a nonempty input string over some alphabet $\Sigma$, whose length $n$ is called the \emph{context length}. Each character of this input alphabet is called a \emph{token} (in practice, tokens are often substrings rather than individual characters; input text is broken up into tokens using a highly nontrivial process called \emph{tokenization}, which is beyond the scope of this survey). The input layer of the transformer embeds each token in a vector space by mapping it to a $d$-dimensional real vector, where $d$ is a parameter of the transformer. Each hidden layer thereafter takes a sequence of length $n$ of $d$-dimensional real vectors as its input, and applies a \emph{length-preserving} function to this sequence, resulting in a  sequence of length $n$ of $d$-dimensional real vectors as the output of that layer. The output layer is different depending on the type of transformer considered.
\begin{itemize}
    \item In a transformer \emph{encoder} (the model adopted in viewing transformers as \emph{classifiers}), the output layer converts the final sequence of $d$-dimensional vectors into a single probability $p_\text{out} \in [0, 1]$, and \emph{accepts} the input string if and only if $p_\text{out} \geq 1/2$.
    \item In a transformer \emph{decoder} (the model adopted in viewing transformers as \emph{language models}), the output layer outputs a new token\footnote{ The new token can be viewed as being drawn from an implicit probability distribution over $\Sigma$. This distribution is effectively learned during training and is encoded in the architecture of the model, thus the model's output procedure  is completely deterministic. In practice, however, the final token is often selected using decoding procedures such as Top-K sampling or Top-p sampling \citep{topp}, which restrict the candidate set before sampling; thus the forward computation is deterministic, while the realized output may be stochastic unless greedy decoding or a fixed random seed is used.}, appends it to the original input, and then continues to do this \emph{autoregressively}, i.e., by sequentially generating new tokens by consuming all the ones generated in previous timesteps, for a pre-specified number of timesteps.
    This is the version of a transformer used for text generation. Furthermore, a decoder can be easily turned into a language recognizer as well: in the final timestep, it behaves similarly to an encoder, outputs a probability $p_\text{out}$ (instead of a new token), and \emph{accepts} the original input string if and only if $p_\text{out} \geq 1/2$.
\end{itemize}



\subsection{Characteristics and Parameters of a Transformer}\label{sec:parameters}

We are now ready to describe the transformer architecture formally. To begin with, every transformer has several characteristics.

\paragraph{Hard/Soft Attention.} The richness of the language of transformers comes from a mechanism inside the hidden layers called \emph{attention} \citep{attentionallyouneed}, which is essentially a scaled dot-product that combines information across different vectors in the sequence. The breakthrough idea behind defining attention was the realization that a model did not need to process language only in a fixed order or compress everything into a single hidden state. Instead, each token could directly ``look at'' the other tokens in a long enough sequence and decide which ones were most relevant. This made models far better at capturing long-range relationships, easier to train in parallel, and scalable to much larger systems --- essentially laying the foundation for modern transformers and large language models. The assumptions on the attention mechanism form a core distinguishing feature of the transformer's behavior. Attention can be either \emph{hard} or \emph{soft}, of which the latter tends to have more expressive power than the former \citep{hahn-2020-theoretical, hao2022formallanguagerecognitionhard, barceloTM, merrillTC0, merrill-etal-2021-effects}. Standard choices for the attention include $\UHAT$ (``unique hard attention''), $\AHAT$ (``average hard attention''), and $\SMAT$ (``softmax attention''), of which, only the last is widely used in practice.

\paragraph{Masking/No Masking.} In an encoder, we typically assume a model without \emph{masking}, which means that every position can \emph{a priori} attend to all other positions. By contrast, in a decoder, we typically assume an autoregressive model that uses \emph{future masking}, where a position can only attend to positions before it.



The evolutions of  LLMs has seen a shift from encoder models (e.g., BERT) to decoder models (e.g., GPT, Claude, Gemini, LLaMA), because of the autoregressive nature of the latter that can be leveraged for text generation. Furthermore, encoder models can be shown theoretically to be strictly more powerful than decoder models\footnote{ It should be noted that these ``less'' powerful decoders do not have \emph{chain-of-thought}, an added capability we describe next.} for language recognition; also, as shown in \citep{CPW24}, lower bounds on encoder models imply lower bounds on constant-depth symmetric circuits; this would be a technical breakthrough, as techniques for circuit lower bound, such as the random restriction method, do not work on symmetric functions.
    
\paragraph{Chain-of-Thought.} In autoregressive decoder models, the architecture can output intermediate tokens during its computation, which are then fed back to the architecture by appending them to the input. This process is called \emph{chain-of-thought} (CoT); it is known that  transformers with this ability are  strictly more powerful than transformers without \citep{CPW24, MS24, LLHZ24}. Most modern LLMs exploit CoT-style intermediate reasoning.

\paragraph{Parameters.} In addition to the preceding characteristics, a transformer  has the following parameters.
\begin{itemize}
    \item \textbf{Number of layers:} the number of hidden layers in the transformer, denoted by $L$. We will always assume $L$ to be a constant, and index the layers as $\ell \in [L]$.
    \item \textbf{Number of attention heads:} the number of attention heads, typically denoted by $H$. We will again  assume that $H$ is a constant, and index the heads as $h \in [H]$.
    \item \textbf{Embedding dimension:} the length of the embedded vectors, denoted by $d$. There are often two additional dimensions, the \emph{key width} $d_\text{key}$ and the \emph{hidden width} $d_\text{hidden}$. Each of these parameters is allowed to depend on the context length $n$, though they are often fixed constants in practice.
    \item \textbf{Level of precision:} the number of bits of precision allowed to carry out all computations within the architecture, denoted by $p$. This parameter is also, in general,  a function of the input length $n$; in fact, we often assume $p$ to be $\Theta(\log n)$ (see the discussion at the start of Section \ref{sec:expressivity}).
    \item \textbf{Amount of chain-of-thought:} for transformers with chain-of-thought, the number of intermediate tokens allowed to be generated, as a function $f(n)$ of the input length $n$. In Section \ref{sec:COT}, we shall see the difference in expressivity resulting from  different asymptotic choices for the function $f(n)$.
\end{itemize}
Sometimes, the number $L$ of layers is called the \emph{depth} of the transformer, while the product $Hdp$ of the embedding dimension, number of attention heads, and number of precision bits used is called its \emph{width}.

It should be emphasized that the context length $n$ (i.e., the length $n$ of an  input)  is \emph{not}  a parameter of a transformer. The reason is that a transformer can process arbitrarily long inputs, the same way a finite automaton can process arbitrarily long strings. This  useful abstraction allows us to view  transformers as language recognizers. In real-world transformers, the context  length, also called the \emph{context window}, is bounded  by some large, but fixed, value (e.g., 256k).

\subsection{Encoder Computation}\label{sec:computation}

\begin{figure}[ht]
\centering
  \includegraphics[scale=0.4]{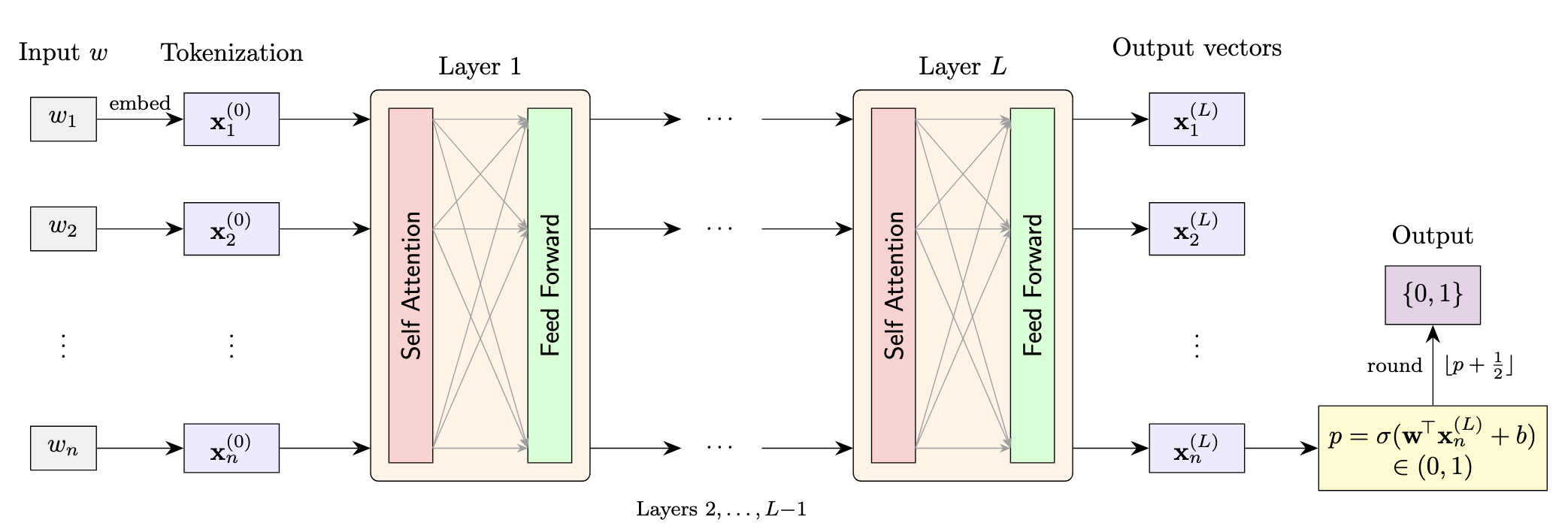}
\caption{A high-level view of the encoder architecture.}
\label{fig:architecture}
\end{figure}

If $X$ is a set, we will write $X^*$ to denote the set of all finite sequences with elements from $X$, while we will write $X^+$ to denote the set of all non-empty such sequences. The set of all real numbers will be denoted by $\mathbb R$. Furthermore, if $m$ is a natural number,  we will write $[m]$ to denote the set $\{1,\ldots,m\}$.




\paragraph{Input Layer.}
In the input layer, a string of length $n$ is mapped to a sequence of $n$ vectors over $\mathbb{R}^d$ via a length-preserving function $\mathsf{embed} : \Sigma^\ast \to (\mathbb{R}^d)^\ast$. To obtain the result of applying the function $\mathsf{embed}$ on a string $w \in \Sigma^\ast$, we take each input character $w_i$ in turn, and take the sum of two functions: the \emph{word embedding} function $\mathsf{WE} : \Sigma \to \mathbb{R}^d$ applied to the character $w_i$, and the \emph{positional encoding} function $\mathsf{PE} : [n] \to \mathbb{R}^d$ applied to the index $i$. The output of the input layer is the resulting sequence $(\mathbf{x}^{(0)}_1, \ldots, \mathbf{x}^{(0)}_n) \in (\mathbb{R}^d)^n$. In other words, we have:
\begin{equation*}
    \mathbf{x}^{(0)}_i = \mathsf{WE}(w_i) + \mathsf{PE}(i), \text{ for all }i \in [n].
\end{equation*}


\paragraph{Hidden Layers.}
Each hidden layer $\ell \in [L]$ of the transformer is a length-preserving function $\mathcal{L}^{(\ell)} : (\mathbb{R}^d)^\ast \to (\mathbb{R}^d)^\ast$ that takes  a sequence $(\mathbf{x}^{(\ell - 1)}_1, \ldots, \mathbf{x}^{(\ell - 1)}_n) \in (\mathbb{R}^d)^n$ as input  and outputs a sequence $(\mathbf{x}^{(\ell)}_1, \ldots, \mathbf{x}^{(\ell)}_n) \in (\mathbb{R}^d)^n$. To describe a hidden layer, we need the notions of  a \emph{self-attention} sublayer and a \emph{position-wise feed-forward} sublayer.



A \emph{self-attention sublayer} with width $d$ and key-width $d_\text{key}$ is a length-preserving function $\mathsf{sa} : (\mathbb{R}^d)^+ \to (\mathbb{R}^d)^+$, in essence the weighted sums of \emph{value} vectors in all $n$ positions, where the weights are a function of \emph{query} vectors and \emph{key} vectors. In other words, we have three matrices $\mathbf{W}^{(Q)}, \mathbf{W}^{(K)}, \mathbf{W}^{(V)} \in \mathbb{R}^{d_\text{key}\times d}$, together with a length-preserving \emph{weighting function} $\mathcal{S} : \mathbb{R}^+ \to \mathbb{R}^+$ and an output matrix $\mathbf{W}^{(O)} \in \mathbb{R}^{d\times d_\text{key}}$, computing the following in an encoder model:
\begin{align*}
    \mathsf{sa}(\mathbf{x}_1, \ldots, \mathbf{x}_n) &= (\mathbf{y}_1, \ldots, \mathbf{y}_n), \text{ where:}
  \end{align*}  
    \begin{align}
    \mathbf{y}_i &= \mathbf{W}^{(O)}\left(\sum_{j = 1}^n\alpha_{i, j}\mathbf{v}_j\right), & 
      \mathbf{v}_j &= \mathbf{W}^{(V)}\mathbf{x}_j, & 
    \alpha_{i, \ast} &= \mathcal{S}(s_{i, \ast}), \label{eq:encoder1}\\
     s_{i, j} &= \frac{\mathbf{q}_i^\top \mathbf{k}_j}{\sqrt{d_\text{key}}},  &
    \mathbf{q}_i &= \mathbf{W}^{(Q)}\mathbf{x}_i, &
    \mathbf{k}_j &= \mathbf{W}^{(K)}\mathbf{x}_j. \label{eq:encoder2}
\end{align}


Here, $s_{i, \ast} := (s_{i, 1}, \ldots, s_{i, n})$ is the vector of \emph{attention scores}, while $\alpha_{i, \ast} := (\alpha_{i, 1}, \ldots, \alpha_{i, n})$ is the vector of \emph{attention weights}.

Note that the weights $\alpha_{i, \ast}$ are obtained by  applying a weighting function $\mathcal{S}$ to the attention scores $s_{i, \ast}$.
The \emph{softmax} function is the most common choice  for a weighting function, where:
\begin{align*}
    [\mathsf{softmax}(a_1, \ldots, a_n)]_i = \frac{\exp(a_i)}{\sum_{j = 1}^n\exp(a_j)}.
\end{align*}
In the literature, several alternatives to softmax have  been considered, such as \emph{hard} attention, where the attention only focuses on the position/s with the maximum score, and either takes one of them (in the $\UHAT$ model), or takes an average over those positions (in the $\AHAT$ model). Since most variants of hard attention can be simulated by softmax attention using positional encodings or other techniques \citep{hardsoftattention}, we focus on the $\SMAT$ model here.

A \emph{position-wise feed-forward sublayer} with width $d$ and hidden width $d_\text{hidden}$ is a function $\mathsf{ff} : \mathbb{R}^d \to \mathbb{R}^d$, in essence a piecewise affine transformation on every position. Thus, we have matrices $\mathbf{W}_1 \in \mathbb{R}^{d_\text{hidden}\times d}$, $\mathbf{W}_2 \in \mathbb{R}^{d\times d_\text{hidden}}$, and vectors $\mathbf{b}_1 \in \mathbb{R}^{d_\text{hidden}}$, $\mathbf{b}_2 \in \mathbb{R}^d$,  so that:
$$
    \mathsf{ff}(\mathbf{x}) = \mathbf{y}, \text{ where }   
    \mathbf{y} := \mathbf{W}_2\mathbf{z} + \mathbf{b}_2 \text{ and } 
    \mathbf{z} := \text{ReLU}(\mathbf{W}_1\mathbf{x} + \mathbf{b}_1).
$$
Here,   the \emph{rectified linear unit} function
$\text{ReLU}(x) = \max(0,x)$  is applied coordinatewise.

Now, for every layer $\ell \in [L]$ and every attention head $h \in [H]$, let $\mathsf{sa}^{(h, \ell)}$ be a self-attention sublayer with width $d$. Similarly, for every layer $\ell \in [L]$, let $\mathsf{ff}^{(\ell)}$ be a feed-forward sublayer with width $d$. The \emph{transformer layer} for layer $\ell \in [L]$ is defined as:
\begin{align*}
    \mathcal{L}^{(\ell)}(\mathbf{x}^{(\ell-1)}_1, \ldots, \mathbf{x}^{(\ell-1)}_n) &= (\mathbf{x}^{(\ell)}_1, \ldots, \mathbf{x}^{(\ell)}_n), \text{ where: }
\end{align*}
\begin{align*}
    (\mathbf{y}^{(\ell)}_1, \ldots, \mathbf{y}^{(\ell)}_n) &:= \sum_{h = 1}^H\mathsf{sa}^{(h, \ell)}(\mathbf{x}^{(\ell-1)}_1, \ldots, \mathbf{x}^{(\ell-1)}_n) + (\mathbf{x}^{(\ell-1)}_1, \ldots, \mathbf{x}^{(\ell-1)}_n), \\
    (\mathbf{x}^{(\ell)}_1, \ldots, \mathbf{x}^{(\ell)}_n) &:= (\mathsf{ff}^{(\ell)}(\mathbf{y}^{(\ell)}_1), \ldots, \mathsf{ff}^{(\ell)}(\mathbf{y}^{(\ell)}_n)) + (\mathbf{y}^{(\ell)}_1, \ldots, \mathbf{y}^{(\ell)}_n).
\end{align*}
While carrying out its computations on an input string $w \in \Sigma^\ast$, the hidden layers of the transformer apply the function $\mathcal{L}^{(\ell)}$ sequentially over the layers $\ell \in [L]$, where the input to layer $1$ is the sequence $\mathsf{embed}(w)$. In effect, therefore, the hidden layers compute the composition:
\begin{align*}
    \mathcal{L}^{(L)}\circ\cdots\circ\mathcal{L}^{(1)}(\mathsf{embed}(w)).
\end{align*}
Note that we have omitted the details of \emph{layer normalization} (or \emph{layernorm} for short), which is a commonly used normalization technique that reduces training time. Layer normalization can change the expressivity of the transformer architecture drastically, depending on how it is modeled; for details, we refer the reader to \citet[Section 6]{cookbook}.

\paragraph{Output Layer.} The last layer $L$ outputs a sequence of length-$d$ vectors $(\mathbf{x}^{(L)}_1, \ldots, \mathbf{x}^{(L)}_n)$. Then, the transformer takes a fixed one of these vectors (typically, $\mathbf{x}^{(L)}_n$), linearly projects it into a scalar, applies a sigmoid function to it to obtain a real number in $(0, 1)$, rounds this number to $0$ or $1$, and outputs the result (interpreted as rejection and acceptance respectively). Thus, we have a vector $\mathbf{w} \in \mathbb{R}^d$ and a scalar $b \in \mathbb{R}$ such that $p = \sigma(\mathbf{w}^\top\cdot
\mathbf{x}^{(L)}_n +b)$,
 where $\sigma$ is the \emph{sigmoid function}, i.e.,
 $\sigma(x) = 1/(1+e^{-x})$. The output of the transformer is $\lfloor p + 1/2\rfloor \in \{0, 1\}$.

 \begin{remark}
     In practice, the weight matrices and vectors throughout the architecture are learned during training and then held fixed at inference time. For expressivity results, it is often useful to impose boundedness assumptions (e.g., on the norm or Lipschitz constant), since such conditions control how  the  output can change under perturbations of the input and rule out pathological behavior. We do not concern ourselves with these considerations in this survey.
 \end{remark}

\subsection{Decoder Computation}

The decoder model is very similar to the encoder model, with the  following two important distinctions.

\paragraph{Masking.}
In encoders, there is no restriction on which positions any particular position can attend to. In decoders, however, each position attends only to the current and previous positions. This is enforced by setting $s_{i, j}=-\infty$, for all $i < j$ in equations \ref{eq:encoder1} and \ref{eq:encoder2} (everything else remains the same). As a consequence of this, all terms with $i < j$ in the expressions vanish. This is called \emph{future masking}. Several other related variants of masking have also  been considered in the literature.

\paragraph{Output Layer.}
In encoders, the output layer projects the vector $\mathbf{x}_n^{(L)}$ into a scalar, and then converts this scalar into a probability. In decoders, the output layer uses $\mathbf{x}_n^{(L)}$ to produce a token from the alphabet $\Sigma$, which can be thought of as the transformer drawing from an implicit probability distribution over the tokens in $\Sigma$. Thus, we have an output function $\gamma : \mathbb{R}^d \to \Sigma$ parameterized as a linear transformation. The output of the transformer is simply $\gamma(\mathbf{x}_n^{(L)})$. 



In decoders with chain-of-thought $f(n)$, the autoregressive nature is leveraged in order to output a sequence of intermediate tokens, for $f(n)$ timesteps. Formally, for a fixed decoder $\mathcal{T}$, let $F_\mathcal{T} : \Sigma^\ast \to \Sigma$ be the function mapping an input string to a token (parameterized by $\mathcal{T}$). For every $w = w_1\ldots w_n \in \Sigma^\ast$, define:
\begin{align*}
    F_\mathcal{T}^0(w) &:= w \\
    F_\mathcal{T}^i(w) &:= F_\mathcal{T}^{i-1}(w)\cdot F_\mathcal{T}(F_\mathcal{T}^{i-1}(w)) \text{ for } i \geq 1,
\end{align*}
where $\cdot$ denotes concatenation. For $j \geq 1$, let $w_{n+j} := F_\mathcal{T}(F_\mathcal{T}^{j-1}(w))$ be the output token  in timestep $j$. Then, the output of the transformer is the sequence of tokens:
\begin{equation*}
    (w_{n+1}, \ldots, w_{n+f(n)}).
\end{equation*}
This transformer can, of course, be easily converted to a language recognizer: instead of generating the final token $w_{n+f(n)}$, the output layer takes the $d$-dimensional vector $\mathbf{x}_{n+f(n)-1}^{(L)}$ and outputs a probability just as an encoder's output layer does, rounding it up or down to represent acceptance or rejection respectively.

Pictorially, the difference between an encoder and a decoder (with CoT $f(n)$) can be visualized as follows:

\begin{figure}[ht]
\centering
  \includegraphics[scale=0.3]{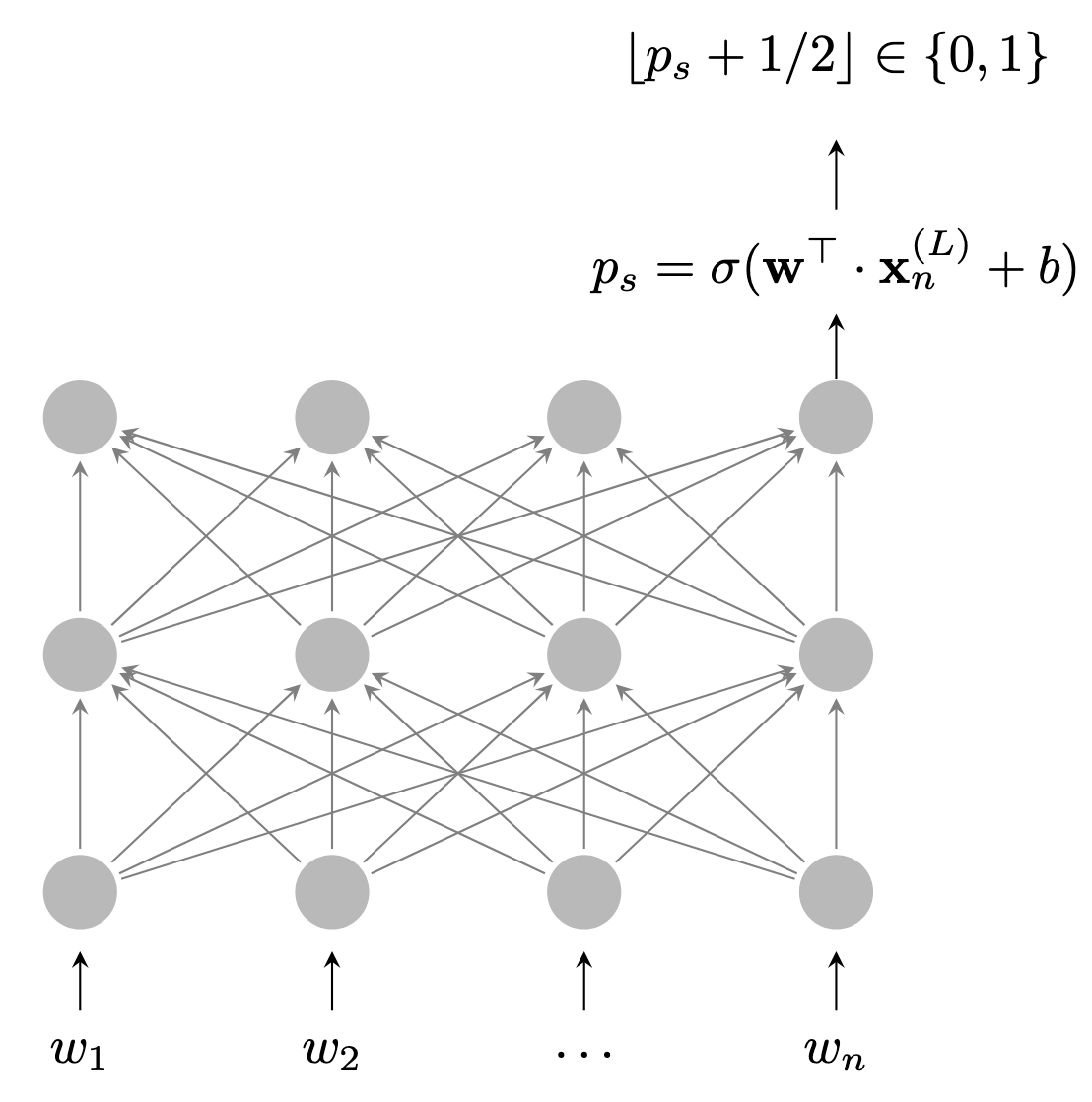}
  \includegraphics[scale=0.3]{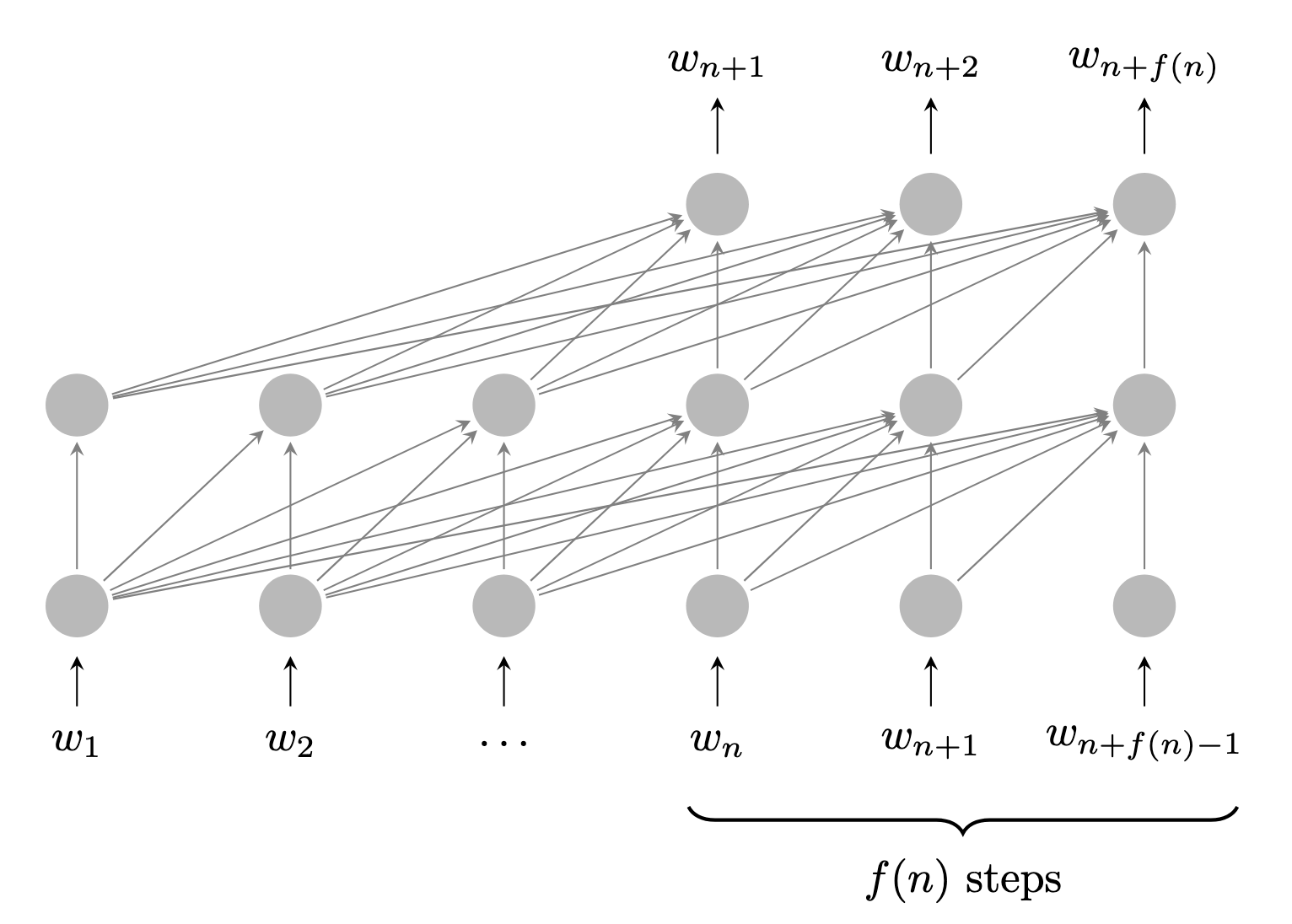}
\caption{An encoder (left) and a decoder (right).}
\label{fig:architectures}
\end{figure}

\section{Classical and Circuit Complexity}\label{sec:complexity}

\subsection{The ``Right'' Hierarchy}\label{sec:hierarchy}

 By and large, circuit complexity has emerged as a particularly well-aligned branch of computational complexity  to calibrate the expressive power of  transformers. The lens of circuit complexity  is more effective than, say, the lens of the Chomsky hierarchy, mainly because the defining inductive bias for transformers is parallel, fixed-depth computation over continuous vectors, rather than discrete symbolic recursion over strings. The Chomsky hierarchy classifies formal languages by the power of grammars or automata, which would be well-suited for models with explicit sequential state transitions (e.g., RNNs). In contrast, transformers operate as layered compositions of attention and feedforward blocks that can be formalized as Boolean or threshold circuits with bounded depth and large fan-in. This alignment is reinforced by empirical and theoretical results.

\subsection{Definitions}\label{sec:definitions}

Formally, a \emph{circuit (on $n$-bit inputs)} is a directed acyclic graph (DAG) $C_n$, whose vertices are called \emph{gates}. A circuit $C_n$ on $n$-bit inputs and size $s$ (for $s > n$) is a DAG on $s$ nodes with some \emph{topological ordering}  $v_1, \ldots, v_s$ of the nodes, i.e., a linear ordering of the nodes such that every node $u$ appears before every node $v$ with an edge from $u$ to $v$. The first $n$  nodes $v_1, \ldots, v_n$ are sources (called the \emph{input gates}), the node $v_s$ is a sink (called the \emph{output gate}), and there are no other sources or sinks. Each gate $v_i$ for $n + 1 \leq i \leq s$ is labeled with a symbol $\sigma_i \in \{\lnot,\land, \lor,  \MAJ\}$. 
The in-degree of every gate labeled $\lnot$ is   $1$, while the in-degree of the other gates can be bigger than $1$. The labels represent standard connectives in Boolean logic with $\MAJ$ being the \emph{majority} function, which evaluates to $1$ if and only if a (strict) majority of its inputs are $1$.

For any $n$-bit input $\mathbf{x} = (x_1, \ldots, x_n)$, the circuit $C_n$ \emph{evaluates}  this input as follows: the \emph{value} of $v_i$ for $1 \leq i \leq n$ is defined to be $x_i$; for each $i \geq n + 1$, the value of $v_i$ is the Boolean function corresponding to the label $\sigma_i$ evaluated on the values of the in-neighbors of $v_i$ (note that all Boolean functions considered here are commutative); the \emph{output} of $C_n$ on input $\mathbf{x}$ is defined as the value of $v_s$. Hence, the circuit $C_n$ can be viewed as a language recognizer over $\{0, 1\}^n$. Stated in other words, $C_n$ \emph{accepts} an $n$-bit input $\mathbf{x}$ if and only if the value of $v_s$ on $\mathbf{x}$ is $1$.

A \emph{circuit family} $\mathcal{C}$ is a sequence $\{C_n\}_{n \in \mathbb{N}}$, where each $C_n$ is a circuit on $n$-bit inputs. Given any $\mathbf{x} \in \{0, 1\}^\ast$, we can choose $C_{|\mathbf{x}|} \in \mathcal{C}$, and evaluate $C_{|\mathbf{x}|}$ on input $\mathbf{x}$ to obtain an output in $\{0, 1\}$. Therefore, each circuit family computes a particular Boolean function $f : \{0, 1\}^\ast \to \{0, 1\}$.

Note that a priori, a circuit family has an arbitrary circuit $C_n$ for each $n \in \mathbb{N}$, but typically we want this family to be presented effectively by some low-complexity function that generates $C_n$ given the value of $n$ in unary. This is the standard notion of circuit \emph{uniformity}. We will only concern ourselves with uniform circuits.

The complexity measures of a circuit family are its \emph{size} (the parameter $s$, which is the number of gates in $C_n$), its \emph{depth} (the length of the largest path from an input gate to an output gate in $C_n$), its \emph{fan-in} (the maximum number of inputs to any gate of $C_n$), and its \emph{basis} (the set of gate labels $\{\sigma_i\}$). The first three of these are functions of $n$. A circuit family is \emph{constant depth} if its depth is a constant independent of $n$. It is \emph{bounded fan-in} if its fan-in is a constant independent of $n$. All circuit families we consider are allowed to have size polynomial in $n$.

Circuit complexity classes are obtained by  constraining how size and depth grow with $n$, and by deciding whether to include the $\MAJ$ label in the basis. We will focus on the following two circuit classes:
\begin{itemize}
    \item $\ACz$: constant-depth, unbounded fan-in, basis $\{\lnot, \land, \lor \}$
    \item $\TCz$: constant-depth, unbounded fan-in, basis $\{\lnot, \land, \lor,  \MAJ\}$.
\end{itemize} It is well-known that:
\begin{equation}\label{chain-of-inclusions}
    \ACz \subsetneq \TCz \subseteq \LOGSPACE \subseteq \POLYTIME,
\end{equation}
where $\LOGSPACE$ is the class of  languages recognized by a Turing machine with a logarithmic number of cells in its work tape and $\POLYTIME$ is the class  of  languages recognized by a Turing machine in polynomial time. 
The first inclusion is strict because the $\MAJ$ function is provably not in $\mathsf{AC}^0$ \citep{FSS};  the next two  inclusions are not known to be strict. In particular, it is open whether $\mathsf{TC}^0 = \POLYTIME$. 



\subsection{Descriptive Complexity}\label{sec:logic}

It is known that  the main computational complexity classes (such as $\POLYTIME$ and $\NP$) and the main circuit complexity classes (such as 
$\ACz$ and $\TCz$) have the same expressive power as certain logical formalisms. 
In particular,  $\ACz$ is equivalent to first-order logic with the BIT predicate, where the $\ACz$-circuits are computed by a random access Turing machine in 
logarithmic time. Furthermore,
$\TCz$ is equivalent to first-order logic with the BIT predicate and ``majority'' quantifiers,  while $\POLYTIME$  is equivalent to least fixed-point logic $\mathsf{LFP}$ on ordered structures.  For a detailed account of the research in this area, which is  known as \emph{descriptive complexity}, see the monograph by
 \citet{neilbook}. 
 
 Results in descriptive complexity have been leveraged in studying the expressivity of transformers. For example, \citet{CCP23} use an extension of first-order logic to show that $\TCz$ contains fixed-precision transformers with softmax attention (\citet{chiang2025transformersuniformtc0} shows that this holds for log-precision transformers as well).
\section{Expressivity Results}\label{sec:expressivity}

In this section, we provide some known expressivity results about transformers, with the corresponding assumptions on the parameters.

However, before proceeding any further, we need to raise the issue of the precision $p$ (see Section \ref{sec:architecture}), which is an important parameter of the transformer architecture. Allowing this precision to arbitrary real numbers can increase the expressivity significantly, but has been widely characterized as unrealistic in practice. On the other hand, limiting the precision to $O(1)$ bits prevents transformers from attending uniformly to length-$n$ strings for growing $n$ \citep{MSlogiclogprecision}; indeed, from a complexity point of view, $O(1)$ bits of precision collapses the expressivity of transformers down to $\ACz$ \citep[Theorem 3.1]{LLHZ24} even with polynomial embedding dimension and $O(\log n)$ steps of chain-of-thought, and the model of computation becomes somewhat less informative for distinguishing transformer variants (see Section \ref{sec:complexity}). A common choice of precision is $\Theta(\log n)$, which is rich enough to allow for addition and rounding conventions.

\subsection{Without Chain-of-Thought}\label{sec:noCOT}

Most expressivity results about transformers \emph{without} chain-of-thought are based on simulation: one fixes a transformer architecture of constant depth and then shows that its computation on a given input can be simulated by an ad hoc circuit family in a low-level circuit class. The relevant circuit class depends strongly on two modeling choices: the type of attention and the amount of numerical precision available as a function of the input length $n$.

Thus, the majority of results in this realm take the form of upper bounds, i.e., they assert that the language recognized by the transformer under consideration is computable by a circuit family of low circuit complexity. The following theorem describes some of the essential containments known, although we encourage the reader to refer to the relevant work for the exact assumptions on the architecture.

\begin{theorem}\label{thm:no-cot}
The following statements are true:
\begin{itemize}
    \item $\UHAT$ encoders with arbitrary (rational) precision only recognize languages in $\ACz$ \citep{hao2022formallanguagerecognitionhard}.
    \item $\SMAT$ and $\AHAT$ encoders with $O(1)$ precision only recognize languages in $\ACz$ \citep{merrillTC0, CCP23, LLHZ24}.
    \item $\SMAT$ and $\AHAT$ encoders with $O(\log n)$-precision only recognize languages in $\TCz$ \citep{merrilluniformTC0, strobltc0, chiang2025transformersuniformtc0}. 
\end{itemize}
\end{theorem}

The basic simulation argument used to prove Theorem \ref{thm:no-cot} is captured by, e.g., \citet[Section 7]{hao2022formallanguagerecognitionhard}, who take an arbitrary encoder with $L$ layers, consider its computation on any fixed arbitrary input, construct small Boolean circuit gadgets to carry out each part of the computation within each transformer layer, and then stitch together these circuit gadgets from different layers. Since $L$ is a constant, this still creates only a constant-depth circuit that simulates the computation of the transformer. $\UHAT$ is weak enough to be simulated only with $\land$, $\lor$, and $\lnot$ gates, and so this process gives rise to an $\ACz$ circuit family. More sophisticated $\AHAT$ or $\SMAT$ machines require the computation of an \emph{average} of $n$ numbers with $O(\log n)$ precision, and this requires threshold gates to compute, resulting in $\TCz$ circuits.

At this juncture, it is reasonable to ask whether or not each containment in Theorem \ref{thm:no-cot} is tight. \citet{barcelo2024logical} show that the result in the first bullet point in Theorem \ref{thm:no-cot} is not tight: there are $\ACz$ languages not recognized by any $\UHAT$ transformers. However, they show that $\UHAT$ transformers do recognize all languages definable in first-order logic with arbitrary unary numerical predicates, which is a rich fragment of $\ACz$. Furthermore, the same paper shows that $\AHAT$ transformers recognize all languages definable in first-order logic with unary numerical predicates \emph{and} counting terms. The results in the second and third bullet points are essentially tight: \citet[Theorems 3.7-3.8]{LLHZ24} show that, when one allows $\mathsf{poly}(n)$ embedding dimension, transformers with $O(1)$ precision and $O(\log n)$ precision capture all of $\ACz$ and $\TCz$, respectively.

There are also several results with a slightly different flavor, utilizing logical characterizations or the Chomsky hierarchy rather than circuit classes. For instance, using an intermediate logic called Boolean RASP (or B-RASP), \citet{YCA24} show that $\UHAT$ decoders (without positional encodings) have the same expressive power as  first-order logic over the natural numbers with the $<$ relation (equivalently, they recognize the class of star-free languages). 


\subsection{With Chain-of-Thought}\label{sec:COT}

Section \ref{sec:noCOT} highlights  that essentially all known results about the expressivity of transformers \emph{without} chain-of-thought tend to put them inside $\TCz$. Chain-of-thought breaks that barrier by going into  classical complexity classes beyond $\TCz$, including $\LOGSPACE$ and $\POLYTIME$, which are believed to be significantly more powerful than $\TCz$ (see   the hierarchy in \eqref{chain-of-inclusions}). This is achieved with appropriate bounds on the chain-of-thought; furthermore, transformers with unbounded chain-of-thought can simulate arbitrary Turing machines.

Some known key results are summarized as follows.

\begin{theorem}\label{thm:cot}
The following statements are true:
\begin{itemize}
    \item $\SMAT$ decoders with $O(\log n)$ CoT and $O(1)$ precision only recognize languages in $\ACz$ \citep{LLHZ24}.
    \item $\SMAT$ decoders with $O(\log n)$ CoT and $O(\log n)$ precision only recognize languages in $\TCz$ \citep{MS24, LLHZ24}.
    \item $\AHAT$ decoders with $O(n)$ CoT and $O(\log n)$ precision only recognize languages in $\mathsf{DTIME}[n^2]$, i.e., deterministic quadratic time \citep{MS24}.
    \item $\AHAT$ decoders with $\mathsf{poly}(n)$ CoT and $O(\log n)$ precision recognize precisely the languages in $\POLYTIME$ \citep{MS24}.
    \item $\AHAT$ decoders with unbounded CoT and arbitrary precision can simulate arbitrary Turing machines \citep{barceloTM, bhattamishra2020, qiuCoT, malach24}.
    \item $\SMAT$ decoders with unbounded CoT and $O(\log n)$ precision can simulate arbitrary Turing machines \citep{jiang2026softmax}.
\end{itemize}
\end{theorem}

The arguments used to prove Theorem \ref{thm:cot} typically involve simulating finite state machines and Turing machines by transformers with chain-of-thought, keeping track of the state and the tape contents by using the generated intermediate tokens, and carrying out each step of the machine computation. Since the contents of the (infinite) tape of the Turing machine cannot be stored in a transformer, the key idea is to encode the computation history by means of the generated tokens. Recognizing the current state of the Turing machine is straightforward to track using the decoder architecture. The difficulty arises in reconstructing the tape symbol being read currently. Roughly speaking, the basic idea leveraged for this is to use the following three steps:
\begin{enumerate}
    \item Use autoregression to compute the sum of the previous head movements, to reconstruct the current head position (using nontrivial techniques such as \emph{layernorm hash} from \citet{MS24});
    \item Find the most recent timestep $t$ when the head was in the same position;
    \item Read off the symbol written on the tape at timestep $t$.
\end{enumerate}

Once again, it is reasonable to ask whether or not   the inclusions in the statement of Theorem \ref{thm:cot} are tight. We have already discussed in Section \ref{sec:noCOT} about the first and second bullet points being near equivalences, for $\mathsf{poly}(n)$ embedding dimension. The third bullet point has a weak partial converse: Every linear-time function is computable by an $\AHAT$ decoder with $O(n)$ chain-of-thought. The fourth and fifth bullet points are equivalences:
every recursively enumerable 
language is computable by a $\UHAT$ decoder with an unbounded amount of chain-of-thought\footnote{Note that this requires a model we have not formalized. Specifically,  all our CoT models require the number $f(n)$ of intermediate tokens to be given in advance; however, transformers that compute arbitrary Turing-recognizable languages do not have $f(n)$ given in advance, but rather have a specific ``acceptance'' token, such that the computation stops and accepts if this token is ever generated. We omit the details here.}. \citet{amiri2025} systematically compute lower bounds on the amount of chain-of-thought required by transformers for various natural algorithmic problems.

\section{Concluding Remarks}\label{sec:discussions}

We gave an overview of the expressive power of transformer models by relating them to circuit complexity classes and logic.

Overall, the complexity-theoretic study of transformer expressivity reveals a nuanced picture: self-attention endows these models with powerful mechanisms for context-dependent computation, yet their abilities depend critically on such resources  as depth, width, precision, positional encoding, and input length. As the field matures, a central challenge is to relate these formal expressivity results to the behavior of real-life trained models, turning insights from worst-case complexity analysis into a sharper understanding of where and why transformers succeed, and where they encounter fundamental limitations.

\section*{Acknowledgments}

We would like to thank Subhash Khot and Andy Yang for very helpful comments on early drafts of this survey.

\bibliographystyle{abbrvnat}
\bibliography{refs}

\end{document}